\documentclass[letterpaper, 10 pt, conference]{ieeeconf}  % Comment this line out if you need a4paper

\IEEEoverridecommandlockouts                              % This command is only needed if 
\usepackage{cite}
\usepackage{amsmath,amssymb,amsfonts}
\usepackage{algorithmic}
\usepackage{graphicx}
\usepackage{textcomp}
\usepackage{xcolor}
\def\BibTeX{{\rm B\kern-.05em{\sc i\kern-.025em b}\kern-.08em
    T\kern-.1667em\lower.7ex\hbox{E}\kern-.125emX}}
    
\usepackage[linesnumbered,ruled]{algorithm2e}
\usepackage{xcolor}
\usepackage{paralist}
\newcommand{\norm}[1]{\left\lVert#1\right\rVert}   
\DeclareMathOperator*{\argmax}{arg\,max}

\usepackage{amsthm}
 
\usepackage[linesnumbered,ruled]{algorithm2e}

\newtheorem{problem}{Problem}

\newtheorem{definition}{Definition}

\usepackage{dirtytalk}
\usepackage[caption=false]{subfig}
\usepackage{mathtools}

\DeclarePairedDelimiter{\abs}{\lvert}{\rvert}
\usepackage{hyperref}
\begin{document}

\title{\LARGE \bf Communication-Aware Multi-robot Coordination with Submodular Maximization
\thanks{\textsuperscript{$\dagger$}Department of Electrical and Computer Engineering, University of Maryland, College Park, MD 20742 USA email:gyshi@terpmail.umd.edu.}

\thanks{\textsuperscript{$\dagger\dagger$}Department of Computer Science, University of Maryland, College Park, MD 20742 USA email: tokekar@umd.edu.}
\thanks{\textsuperscript{$\ddagger$}GRASP Laboratory,
University of Pennsylvania, Philadelphia, PA, USA e-mail:
lfzhou@seas.upenn.edu. The author was with the Department of Electrical and Computer Engineering, Virginia Tech, Blacksburg, VA, USA when part of the work was completed.}
}

\author{{Guangyao Shi\textsuperscript{$\dagger$}},
{Ishat E Rabban\textsuperscript{$\dagger\dagger$}},
{Lifeng Zhou\textsuperscript{$\ddagger$}}, 
{Pratap Tokekar\textsuperscript{$\dagger\dagger$}}\\
}

\maketitle
\begin{abstract}
Submodular maximization has been widely used in many multi-robot task planning problems including information gathering, exploration, and target tracking. However, the interplay between submodular maximization and communication is rarely explored in the multi-robot setting. In many cases, maximizing the submodular objective may drive the robots in a way so as to disconnect the communication network.  Driven by such observations, in this paper, we consider the problem of maximizing submodular function with connectivity constraints. Specifically, we propose a problem called Communication-aware Submodular Maximization (CSM), in which communication maintenance and submodular maximization are jointly considered in the decision-making process. One heuristic algorithm that consists of two stages, i.e. \textit{topology generation} and \textit{deviation minimization} is proposed. We validate the formulation and algorithm through numerical simulation. We find that our algorithm on average suffers only slightly performance decrease compared to the pure greedy strategy. 
\end{abstract}

\iffalse
\fi
 \section{Introduction}
 Many multi-robot cooperative tasks such as information gathering, exploration, target tracking can be formulated as submodular maximization problems \cite{zhou2018resilient,corah2019distributed,williams2017decentralized,jorgensen2017matroid,dames2017detecting}. The objectives in such problems (e.g., mutual information, area explored, number of targets tracked, etc.) have diminishing returns property which is shown to be submodular. Intuitively, submodularity formalizes the notion that adding more robots to a larger multi-robot team cannot yield a larger marginal gain in the objective than adding the same robot to a smaller team. Despite the fact that the submodular maximization problem is NP-hard, a simple greedy approximation algorithm can achieve nearly optimal performance \cite{nemhauser1978analysis,krause2011submodularity}.  
 
 Communication plays a key role in successfully executing the greedy algorithm for a team of robots to choose actions. In the centralized case, robots may need to transmit their acquired information to either a leader of the team or a remote server; in the distributed case, robots may need to exchange local information with their neighbors to reach some global consensus \cite{alonso2019distributed,cristofalo2019consensus}. In either case, a team of robots needs to form a connected communication network for ensuring information flow. However, it is possible that actions that maximize the objective may lead the network to become disconnected. In fact, we expect that to happen quite often given that many of these objectives relate to coverage where the robots may want to move away from each other to reduce overlaps. Therefore, there is a need to introduce a connectivity constraint during decision-making. However, in the existing literature related to submodular maximization, communication is usually neglected, which motivates us to think about whether we can jointly consider communication maintenance and submodular maximization in the decision-making process for the team.

\begin{figure}
    \centering
    \includegraphics[scale=1]{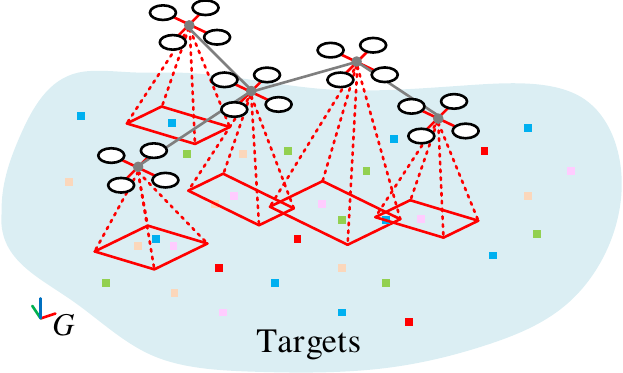}
    \caption{One motivating example of this paper: multi-robot active target tracking with communication constraints. The team should maximize the number of targets observed at each planning step and and keep connected. Colored squares represent targets. Grey dots and lines represent induced communication network.}
    \label{fig:outline}
\end{figure}

To this end, we propose a problem named Communication-aware Submodular Maximization (CSM) for a class of multi-robot task planning. In the proposed formulation, each robot needs to find one trajectory to be executed within the current planning epoch to maximize the submodular team objective. We also impose the constraint that the trajectories that are found must be such that the robots form a connected communication network at the end of the planning epoch. That is, the end positions of each of the trajectories that must form a connected network. We allow the robots to temporarily disconnect during the epoch. However, by ensuring connectivity at the end of the epoch, the robots will be able to exchange information gained during the epoch and jointly plan for the next epoch. 

One motivating example is given in Fig. \ref{fig:outline}, in which a team of aerial robots equipped with downward-facing cameras are tracking targets on the ground. The objective is to maximize the number the targets observed within each epoch. On the one hand, robots need to move away from each other to reduce the overlap of the sensor footprints. On the other hand, they cannot move too far from each other while still ensuring a connected network at the end. A good solution for CSM is able to balance these two conflicting goals.
In this paper, we propose a heuristic algorithm consisting of two stages, \textit{topology generation} and \textit{deviation minimization}, to solve CSM. The key idea of the proposed algorithm is that first, for each robot we discretize the problem by generating a set of candidate trajectories whose endpoints are within the reachable set of the robot and let robots choose trajectories greedily without considering communication constraints; then we make them minimally deviate from the endpoints corresponding to greedy selections to build connectivity. Specifically, in the \textit{topology generation} stage, an edge-weighted graph is generated using robots' greedy selections, whose edge weights are defined over distance between pairs of robots. Then a Minimum Spanning Tree (MST) of the edge-weighted graph is extracted as the communication graph for the next epoch.
In the \textit{deviation minimization} stage, a Quadratic Programming (QP) is formulated to find new positions within the reachable set that minimally deviate from the greedy selections for robots to realize the MST in the workspace. We carry out extensive simulations to evaluate the performance of the proposed algorithm.

\section{Related Work}
Submodular maximization and its variants have been widely used in multi-robot decision-making problems including coverage that \cite{sun2019exploiting,ramachandran2019resilient,krause2008near}, target tracking \cite{zhou2018resilient, zhou2019sensor,zhou2020distributed,tokekar2014multi}, exploration \cite{corah2019distributed}, and information gathering \cite{schlotfeldt2018resilient,shi2020robust,jorgensen2017matroid,krause2011submodularity,krause2006near}. These studies all based on the fact that greedy algorithm and its variants can solve submodular maximization problem its variants efficiently with provable performance guarantee. a However, communication is seldom considered in the existing works from robotic researchers and most of them assume that conna ected network is always there. Krause \cite{krause2006near} uses the same assumption but associates some cost to edges of network and jointly optimizes coverage and network costs. But their work is static in nature and decision is made only once instead of at each planning epoch without partition matroid constraint \cite{schlotfeldt2018resilient}. By contrast, we consider the scenario where robots are moving and decisions need to be made at each planning epoch to maximize objective and guarantee connected network. 
Gharesifard \cite{gharesifard2017distributed} considers the submodular maximization case where one decision-maker represented as one vertex in the graph can access only the decisions of its neighbor and analyzes the influence of graph topology. \cite{grimsman2018impact} considers a similar problem but mainly from the perspective of a system designer. However, the graph discussed in \cite{gharesifard2017distributed,grimsman2018impact} is more like a relational graph instead of a communication graph, and authors all assume that the actions or decisions of vertices will not change the graph properties, i.e.,  connectivity. Grimsman \cite{williams2017decentralized} considers submodular maximization under topology constraints for multi-robot task allocation problems. If the topology constraints can be described as matroid constraints, a greedy algorithm can have a constant approximation factor. However, connectivity constraint is in general not a matroid constraint for robot team with the same communication radius, and authors in \cite{williams2017decentralized} make extra assumptions on task allocation structures to make spanning tree constraint a matroid constraint, which makes it unsuitable for our problem.  

Another line of related work is on connectivity maintenance based on algebraic graph theory \cite{sabattini2013decentralized,zavlanos2007flocking,zavlanos2011graph,ji2007distributed}. In these studies, connectivity maintenance and task are usually separately considered. Even though connectivity can be formally guaranteed when the task is not considered, complex task behaviors of robots in practice can break the connectivity. Our work is different from these works in two aspects: first, our formulation is rooted in combinatorial optimization, i.e., submodular maximization rather than continuous optimization and aims at solving discrete decision-making problems. Even though submodular set functions can be extended to continuous functions \cite{bach2011learning}, the evaluation of the extended function involves exponentially many subsets of the ground set, which makes it unsuitable for robotic application. Second, we jointly consider team task and communication connectivity and therefore task behaviors will not break connectivity.

This work is also closely related to submodular maximization over graphs \cite{huang2019maximizing,kuo2014maximizing,ghuge2020quasi}. \cite{ghuge2020quasi} considers the problem of finding a rooted arborescence in a directed graph with a budget to maximize a submodular function while by contrast graph considered in this paper is undirected and budget is not a constraint. In \cite{huang2019maximizing}, authors consider the problem of finding a connected subgraph to maximize covered area, which is a special submodular set function, in the Euclidean plane and give one $2$-approximation the algorithm. But their algorithm cannot be used to maximize general submodular functions that are interesting to multi-robot applications, for example mutual information \cite{dames2017detecting,corah2019distributed} and number of targets \cite{zhou2018resilient} because the performance of their algorithm relies on some additive properties of covered areas. \cite{kuo2014maximizing} considers a similar problem as with \cite{huang2019maximizing} for general submodular functions. However, none of algorithms proposed in  \cite{huang2019maximizing,kuo2014maximizing,ghuge2020quasi} can deal with partition matroid constraints \cite{corah2019distributed}, which is one inherent nature in multi-robot application since each robot has its unique set of choices. As a result, corresponding algorithms are not applicable to multi-robot applications. 

\textbf{Contribution}:
the main contribution of this paper is that we formulate a novel submodular maximization problem with connectivity constraints for multi-robot applications. A heuristic algorithm consisting of two stages is proposed. The correctness of the formulation and the proposed algorithm is validated through a case study on active target tracking. 

\section{Problem Formulation}
\subsection{Preliminary}
We first introduce some notations we will use in this paper.
 Given a set $\mathcal{A}$, $2^{\mathcal{A}}$ denotes the power set of $\mathcal{A}$. Given another set $\mathcal{B}$, the set $\mathcal{A} \setminus \mathcal{B}$ denotes the set of elements in $\mathcal{A}$ but not in $\mathcal{B}$. Given a set $\mathcal{V}$, a set function $f: 2^{\mathcal{V}} \mapsto \mathbb{R}$, and an element $v \in \mathcal{V}$, $f(v)$ is a shorthand that denotes $f(\{v\})$. We use $\Delta_f(r \mid  \mathcal{A})=f(\mathcal{A} \cup \{r\})-f(r)$ to denote the marginal gain of adding $r$ in $\mathcal{A}$ .
Whenever $f$ is clear from the context, we will use shorthand $\Delta(r \mid  \mathcal{A})$ for $\Delta_f(r \mid  \mathcal{A})$.
 $[N]$ denotes the set $\{1,2,\ldots,N\}$.

\begin{definition}[Minimum Spanning Tree]
A minimum spanning tree (MST) is a subset of the edges of a connected, edge-weighted undirected graph that connects all the vertices together, without any cycles and with the minimum possible total edge weight.
\end{definition}

\begin{definition}[Minimum Bottleneck Spanning Tree]
A minimum bottleneck spanning tree (MBST) in an undirected graph is a spanning tree in which the most expensive edge is as cheap as possible.
\end{definition}
\subsection{System model}
Suppose we have $N \geq 2$ robots in the environment, whose joint states at the start of the current epoch are denoted as $\mathbf{x}=\{x_1,x_2,\ldots,x_N\} \in \mathbb{R}^{dN}$,where $d$ is the dimensionality and can be 2 or 3. Communication graph at the start of the current epoch $G=({V}, {E})$ is specified using proximity graph, i.e., a robot corresponds to a vertex $i \in {V}$ and $(i,j) \in {E}$ if $\norm{x_i-x_j} \leq r_c$, where $r_c$ is the communication radius.

Each robot is equipped with a sensor that gathers information based on the sensor footprint. For example, an aerial robot equipped with a downward-facing camera can sense the targets on the ground as shown in Fig. \ref{fig:outline}. In this paper, specifically, our goal is to find a set of trajectories, one per robot, to ensure that we maximize the submodular team objective within each epoch and ensure a connected network at the end of each epoch.

\iffalse
 Let $\mathcal{M}=\{m_1,m_2,\ldots,m_M\}$ be a set of macro actions for robots. For example, $m_i$ may stand for going north for some distance.
 We assume that some lower-level motion control algorithms can achieve such movements.
 We denote $\mathcal{A}_i=\{(i,m_1),(i,m_2),\ldots,(i,m_M)\}$ as the action set for robot $i$ and $\mathcal{G}=\cup_{i=1}^N \mathcal{A}_i$ as the ground set for actions for all robots. We define the partition matroid as $(\cup_{i=1}^N \mathcal{A}_i, \mathcal{I})$, where $\mathcal{I}=\{{\mathcal{A}} \in 2^{\mathcal{G}} \mid \lvert {\mathcal{A}} \cap \mathcal{A}_i \rvert \leq 1,  1 \leq i \leq N\}$.
 \fi
 Let $\mathcal{R}_i$ be the reachable set of robot $i$ of the current epoch, i.e., a set of locations that robot $i$ can reach at the end of the current epoch from the start position of the current epoch. In general, there is no closed-form expression for the reachable set. In this paper, we assume that the reachable set can be approximately represented as obstacle-free convex polyhedrons or ellipsoid using an iterative optimization method such as the one presented in \cite{deits2015computing}. Let $\mathcal{T}_i$ be the set of trajectories of robot $i$ each of which is within the reachable set $\mathcal{R}_i$. It should be noted that there are infinitely many elements in $\mathcal{T}_i$. Let $\mathcal{P}$ be the operator that can extract the endpoints of trajectories. With the above notations, the problem can be formulated as follows. 

 \begin{problem}[Communication-aware Submodular Maximization]\label{problem:SMIC}
 At current planning epoch, robots' positions induce a connected communication graph $G_e$, we want to preserve the connectivity in the next epoch, i.e., induced graph $G_{e+1}$ after moving to new positions is connected, meanwhile maximizing a submodular objective function. Mathematically, 
 \begin{equation}
     \begin{aligned}
     \max_{s_i \in \mathcal{T}_i, i \in [N]}~& f(\mathcal{S}) \\
     s.t.~&~ G_{e+1}(\mathcal{P}(\mathcal{S}))~\text{is}~\text{connected,}
     \end{aligned}
 \end{equation}
 where $\mathcal{S} = \{s_1, \ldots, s_N\}$ and $\mathcal{T}_i$ is the set of trajectories for robot $i$.
 \end{problem}

One way to solve Problem \ref{problem:SMIC}is by first discretizing the reachable set $\mathcal{R}_i$ into a set of discrete locations and by using lower-level motion primitives such that each point in the discretized reachability set is associated with one dynamically feasible trajectory \cite{branicky2008path}.  

Even after discretization,  Problem \ref{problem:SMIC} is still hard because the graph topology can be quite diverse and there can be exponentially many candidates. However,   
 it should be noted that with proper discretization there is always a solution to Problem \ref{problem:SMIC}: since communication graph $G_{e}$ is connected and all robots have non-empty reachable sets, one apparently feasible solution is that all robots do the same movement, e.g., the team as a whole shifts to a new position and $G_{e+1}$ has the same graph topology as that of $G_{e}$. 

 \begin{figure}
    \centering
    \includegraphics[scale=0.95]{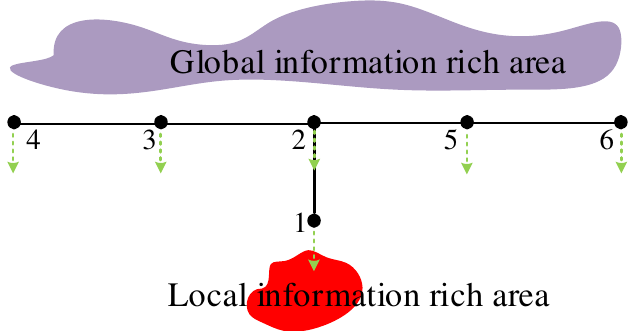}
    \caption{One counter example to show that the sequential graph greedy strategy can be arbitrarily bad. If robot 1 is attracted to some local information rich area and firstly chooses an action to go downwards (dotted blue arrow), it may bias the whole team away from the global information rich area, which is on the top (other robots are forced to choose the blue arrows for connectivity).}
    \label{fig:greedy_bad_example}
\end{figure}
 
One naive way to solve Problem \ref{problem:SMIC} is to use a sequential greedy strategy and consider connectivity during the construction process as shown in Algorithm \ref{algorithm:sequential greedy}, in which at each iteration one robot that can connect to the existing connected graph will make decisions and the connected graph will then be correspondingly expanded. However, such a strategy can be arbitrarily bad as shown in Fig. \ref{fig:greedy_bad_example}. In the Sequential Graph Greedy (SGG) strategy, robot 1 may choose the action to go down first since it has the largest marginal gain w.r.t. empty graph, but his action will bias the team from more valuable area on the top. Another problem with such a strategy is that if robots have different reachable sets, SGG may end up with a disconnected graph. For example, in Fig. \ref{fig:greedy_bad_example}, if robot 1 has a large reachable set and move downwards to its best, other robots with a smaller reachable set may not be able to follow its pace to maintain the connectivity. 

\begin{algorithm}[ht]\label{algorithm:sequential greedy}
    \caption{Sequential Graph Greedy}
    \SetKwInOut{Input}{Input}
    \SetKwInOut{Output}{Output}
    \underline{function SGG}($f,\{\mathcal{T}_i\}_{i=1}^{N},\mathcal{P}$) \\
    \Input{
    \begin{itemize}
        \item A monotone submodular function $f$
        \item Partitioned ground set $\{\mathcal{T}_i\}_{i=1}^{N}$
        \item Operator $\mathcal{P}$ that  can  extract  the  end  points of  trajectories.
    \end{itemize}
    }
    \Output{
    A subset $Sol$ of the ground set
    }
    $Sol \gets \emptyset$,
    $\mathcal{T} \gets \bigcup_{i=1}^{N} \mathcal{T}_i$,
    $G \gets (V = \emptyset, E=\emptyset)$\\
    \While{$\abs{Sol} < N$}{
    find the trajectory set $\mathcal{T}^{\prime} \subseteq \mathcal{T}$ s.t. each element in $\mathcal{T}^{\prime}$ can establish connections to the existing graph $G$ at its endpoint\\
    
    \# find the element with largest marginal gain and its group ID \\
    $a,~ i = \argmax_{s \in \mathcal{T}^{\prime}}\Delta f (s \mid Sol)$ \\
    $\mathcal{T} \gets \mathcal{T} \setminus \mathcal{T}_i$,
    ~~$Sol \gets Sol \cup \{s\}$ \\
    G.add\_node($\mathcal{P}(s)$),~~ 
    G.add\_edge($\mathcal{P}(s)$)
    }
    return $Sol$
\end{algorithm} 

\begin{algorithm}[ht]\label{algorithm:matroid}
    \caption{Submodular Maximization with Partition Matroid}
    \SetKwInOut{Input}{Input}
    \SetKwInOut{Output}{Output}
    \underline{function Greedy}($f,\{\mathcal{T}_i\}_{i=1}^{N}, \mathcal{P}$) \\
    \Input{
    \begin{itemize}
        \item A monotone submodular function $f$
        \item Partitioned ground set $\{\mathcal{T}_i\}_{i=1}^{K}$
        \item Operator $\mathcal{P}$ that  can  extract  the  end  points of  trajectories
    \end{itemize}
    }
    \Output{
    A subset $Sol$ of the ground set
    }
    $Sol \gets \emptyset$,
    $\mathcal{T} \gets \bigcup_{i=1}^{N} \mathcal{T}_i$ \\
    \While{$\abs{Sol} < N$}{
    \# find the element with largest marginal gain and its ID \\
    $s,~ i = \argmax_{s \in \mathcal{T}}\Delta (s \mid Sol)$ \\
    $\mathcal{T} \gets \mathcal{T} \setminus \mathcal{T}_i$ \\
    $Sol \gets Sol \cup \{s\}$
    }
    return $Sol$
\end{algorithm}

 \begin{figure*}
    \centering
    \includegraphics[scale=0.92]{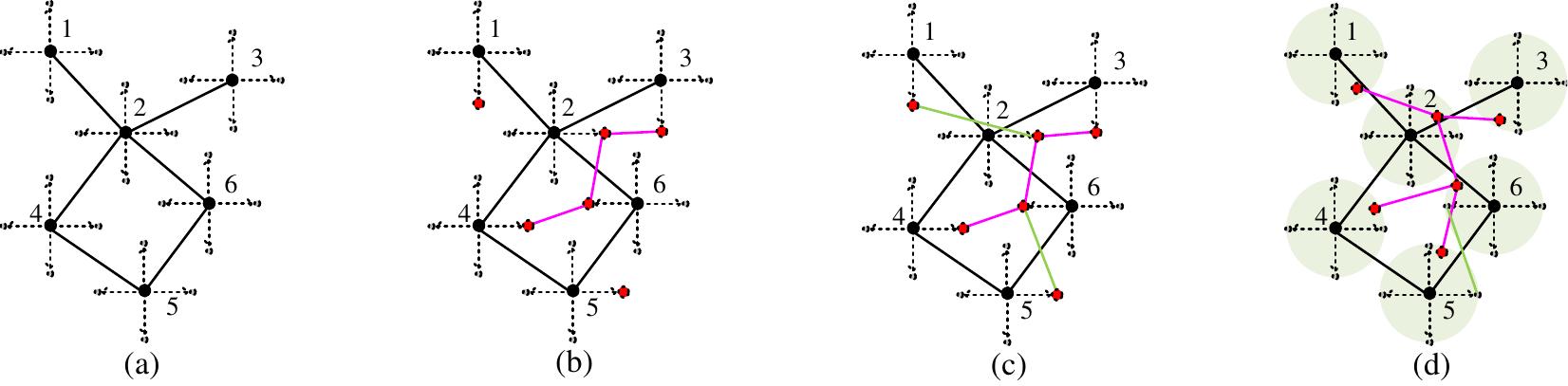}
    \caption{One illustrative example of the proposed algorithm. (a) There are $N=6$ robots in the team and each robot has five trajectories: go left, go right, stay, go up, and go down which corresponds to five dotted line in the work space. Current position of robots are denoted as black dots and the corresponding communication links are black lines. (b) The team greedily maximizes the submodular objective function which results in a disconnected graph i.e. robots 2, 3, 4, 6 are connected but robot 1 and 5 are disconnected from other robots. (c) Generate a complete weighted graph using greedy selections and then extract a MST from weighted graph $K_6$. Two blue edges are added. (d) Solve the deviation minimization problem to find the final positions for robots. The shaded circles centered at current positions are reachable sets for robots and the red graph is the network to be realized at next epoch.} 
    \label{fig:algorithm}
\end{figure*}

\section{Heuristic Algorithm} 
We propose a two-stage heuristic algorithm to solve Problem \ref{problem:SMIC}. The key idea is to first let robots greedily select trajectories, which of course may break the connectivity, and then make the robots minimally deviate from greedy selections to establish connectivity.

\textbf{Topology  Generation}: In this stage, robots will greedily select trajectories based on marginal gains without considering the communication constraint. It is equivalent to solving a submodular maximization problem with partition matroid using a greedy strategy as shown in Algorithm \ref{algorithm:matroid}. These greedy selections will virtually drive them to positions ${x_1^g,\ldots,x_N^g}$. Then we generate a complete graph $K_N$ corresponding to these $N$ robots. The edge weight between robots $i$ and $j$ is defined as $C(i,j)=\max(\frac{1}{2}(\norm{x_i-x_j}-r_c),0)$. If the distance between two robots is already within the communication radius, then the cost of that edge will be zero. If the distance is greater than $r_c$, then at least one of the two robots needs to move at least $\frac{1}{2}(\norm{x_i-x_j}-r_c)$ to realize the edge. 

After defining the $K_N$, we use MST algorithm \cite{cormen2009introduction} to extract a connected graph for the next epoch. The intuition for using MST to generate graph topology is: first, the costs of edges are defined over distances to be traveled to realize edges. Therefore, the graph returned by MST algorithm to some extent reflects the total deviation to achieve connectivity; second, as shown in \cite{cormen2009introduction}, MST is also a MBST, which suggests that the graph returned by MST algorithm to some extent also reflects the minimum distance to be deviated to achieve connectivity.

\textbf{Deviation Minimization}: After generating the graph topology, in this stage we consider the problem of how to make robots minimally deviate from the greedy selections to establish connectivity. The problem is described below.
\begin{problem}[Deviation Minimization]\label{problem:deviation minimization}
Given current positions ${x_1,\ldots,x_N}$ of robots, positions ${x_1^g,\ldots,x_N^g}$ corresponding to greedy selection, reachable sets $\mathcal{R}_1, \ldots, \mathcal{R}_N$, communication radius $r_c$, and a connected graph $G_{e+1}$ to be realized, the goal is find new locations ${x_1^*,\ldots,x_N^*}$ such that 
\begin{itemize}
    \item ${x_1^*,\ldots,x_N^*}$ can realize all edges in graph $G_{T}$, which is returned by MST algorithm;
    \item The deviations of robots from greedily selected positions ${x_1^g,\ldots,x_N^g}$ are minimized.
    \item ${x_1^*,\ldots,x_n^*}$ are within reachable set w.r.t their current
    positions.
    \item Robots will not collide in the new positions.
\end{itemize}
Mathematically, 
\begin{equation}
    \begin{aligned}
    \min \sum_{i=1}^{N} & w_i\norm{x_i^*-x_i^g} \\
     s.t. 
    & \norm{x_i^*-x_j^*} \leq r_c, (i,j) \in G_T.edges \\
    & x_i^* \in \mathcal{R}_i, i \in [N] \\
    \quad &\norm{x_i^*-x_j^*} > r_s, \forall i \neq j ,
    \end{aligned}
\end{equation}
where $w_i \geq 0$ is parameter used to describe robot $i$'s willingness to deviate from $x_i^g$ and more details on choosing $w_i$ are given in Sec. \ref{sec:case study} and $r_s$ is the safe radius for each robot.
\end{problem}
It should be noted that the safety constraints are not convex but they can be well handled by available solves like Gurobi 9.0 \cite{gurobi}.

One illustrative example is shown in Fig. \ref{fig:algorithm} to demonstrate how our proposed algorithm works. In the formulation given above, the reachable set $\mathcal{R}_i$ can be an arbitrary set. We also have not defined $w_i$ which can be chosen based on the application. In the following, we will describe one specific example on active target tracking in which we assume that the reachable set can be represented as a circle and present three ways to select $w_i$ based on the number of targets that a robot can track during one epoch.

\iffalse
\begin{algorithm}[ht]\label{algorithm:matroid}
    \caption{Submodular Maximization Partition Matroid}
    \SetKwInOut{Input}{Input}
    \SetKwInOut{Output}{Output}
    \underline{function Greedy}($f,\{\mathcal{X}_i\}_{i=1}^{K},k$) \\
    \Input{
    \begin{itemize}
        \item A monotone submodular function $f$
        \item Partitioned ground set $\{\mathcal{X}_i\}_{i=1}^{K}$
    \end{itemize}
    }
    \Output{
    A subset $Sol$ of the ground set
    }
    $Sol \gets \emptyset$,
    $\mathcal{X} \gets \bigcup_{i=1}^{K} \mathcal{X}_i$ \\
    \While{$\abs{Sol} < K$}{
    \# find the element with largest marginal gain and its group ID \\
    $s,~ i = \argmax_{s \in \mathcal{X}}\Delta f (s \mid Sol)$ \\
    $\mathcal{X} \gets \mathcal{X} \setminus \mathcal{X}_i$,
    $Sol \gets Sol \cup \{s\}$
    }
    return $Sol$
\end{algorithm}
\fi

\section{Case Study: Active Target Tracking}\label{sec:case study}
In this section, we present one case study on multi-robot active target tracking with  communication constraints, in which each robot has a downward-facing camera and the team aims to maximize the number of targets that will be observed in each planning epoch. This case study is presented to demonstrate: 
\begin{itemize}
    \item Correctness: The proposed algorithm will return a solution that preserves connectivity. This will be validated by checking the network generated by the team after taking action.
    \item Performance: The proposed algorithm outperforms SGG strategy and has a competitive performance compared with a simple greedy strategy, which can be viewed as the empirical performance upper bound for the team at each planning epoch. We compare our algorithm with these two algorithms with respect to the number of targets tracked.
\end{itemize}

\subsection{System Model}
\textbf{Target Model}: We assume each target has single integrator motion model,
\begin{equation*}
    p^j(t+1) = p^j(t) + v^j(t),
\end{equation*}
where $p^j(t)$ and $v^j(t)$ denote the position and the velocity of target $j=1,\ldots,50$. The robot obtains noisy measurements of the targets' positions and use a Kalman filter for estimation. The noise is set to Gaussian noise with zero mean and 0.5 standard deviation. Each target's velocity is initialized to be zero and is updated by using two consecutive measurements and the time interval of these two measures:
\begin{equation*}
    v^j(t^{\prime}) =\frac{\Tilde{p}^j(t^{\prime})-\Tilde{p}^j(t)}{t^{\prime}-t},
\end{equation*}
where $\Tilde{p}^j(t^{\prime})$ and $\Tilde{p}^j(t)$ are two measures at time step $t^{\prime}$ and $t$ with $t^{\prime}>t$.

 \begin{figure}
    \centering
    \includegraphics[scale=1]{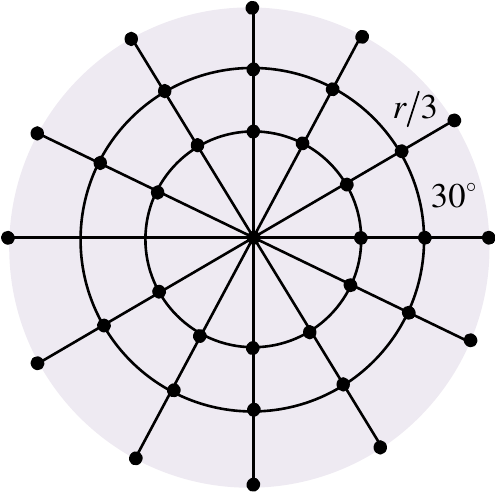}
    \caption{One illustrative example to demonstrate the discretization of the reachable set. Endpoints are uniformly sampled w.r.t radius (step size $\frac{r}{3}$) and angles (step size $30^{\circ}$)}.
    \label{fig:discretization}
\end{figure}
\textbf{Robot Model}: We assume that all aerial robots fly at fixed heights and the reachable set of each robot can be described as a circle centered at its current position with a radius $4$ meters. Then the reachable set is discretized w.r.t. both the radius and the angle as shown in Fig. \ref{fig:discretization} to generate trajectory set. Each robot is equipped with a downward-facing camera and is able to observe ground targets inside the sensor footprints. We conduct 10 rounds of simulation and in each simulation there are 10 epochs. Communication radius is set to be 10 meters.

\textbf{Weight Selection}: in this paper we test two ways to set parameter $w_i$, which is about robot $i$'s willingness to deviate from its endpoints, for Problem \ref{problem:deviation minimization}. 
\begin{figure*}[ht]
    \subfloat[\label{fig:animation 1}]{
    \includegraphics[width=0.23\textwidth]{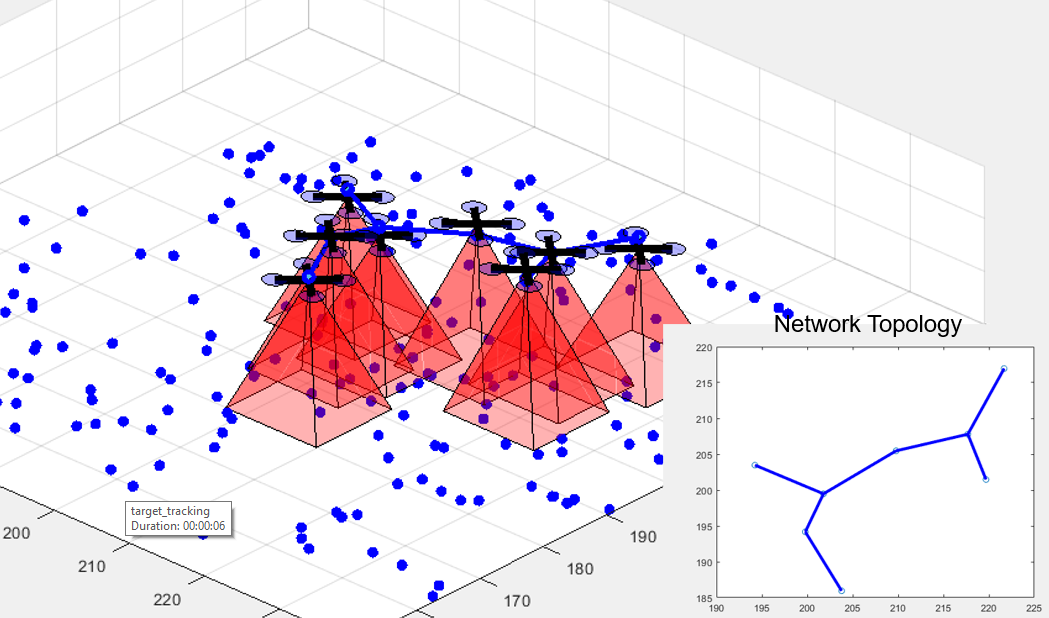}
    }
    \subfloat[\label{fig:animation 2}]{
    \includegraphics[width=0.23\textwidth]{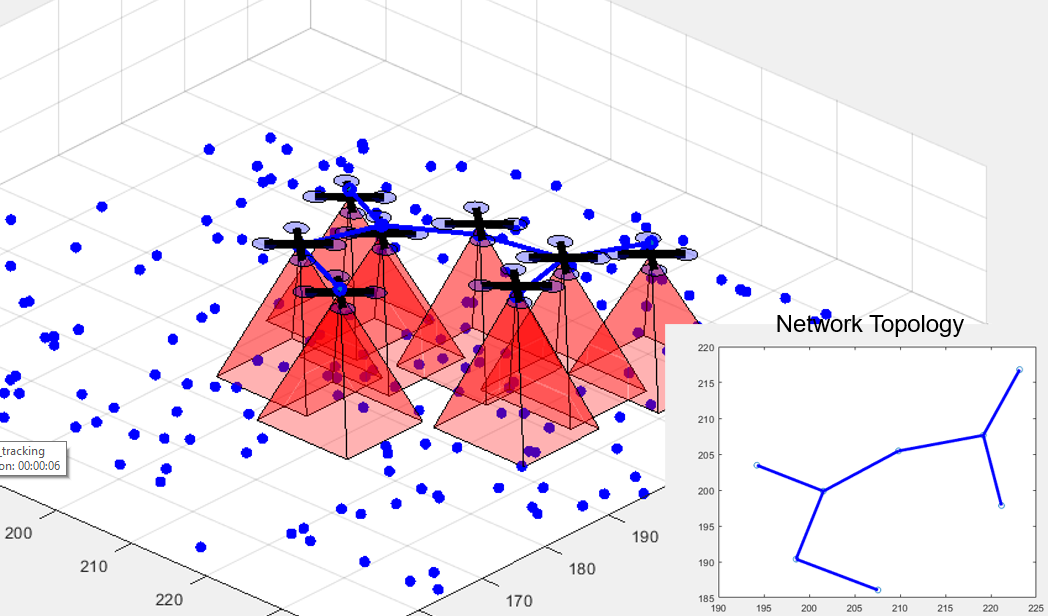}
    }
    \subfloat[\label{fig:animation 3}]{
    \includegraphics[width=0.23\textwidth]{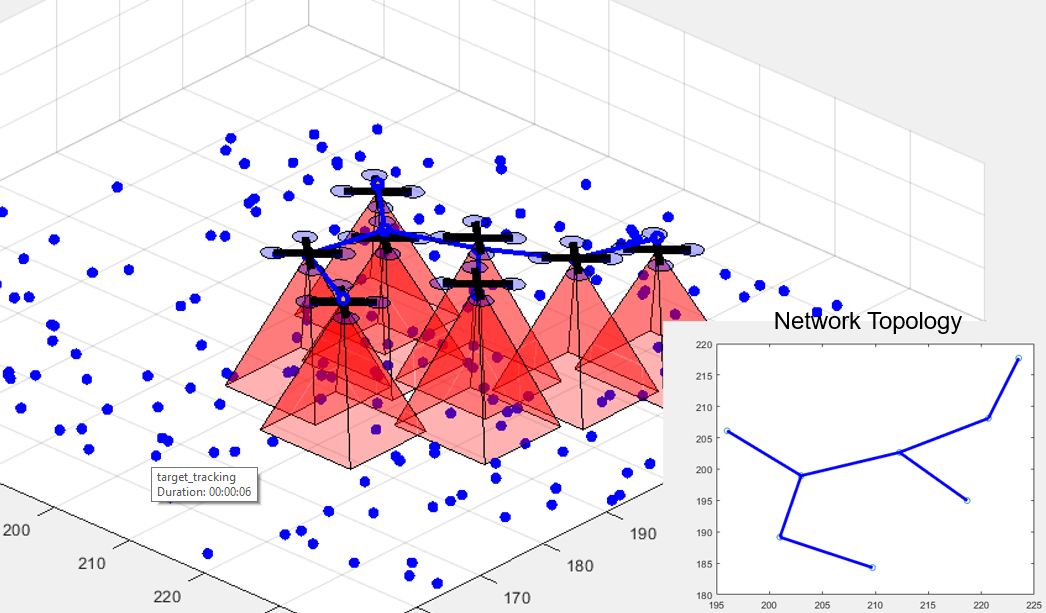}
    }
    \subfloat[\label{fig:animation 4}]{
    \includegraphics[width=0.250\textwidth]{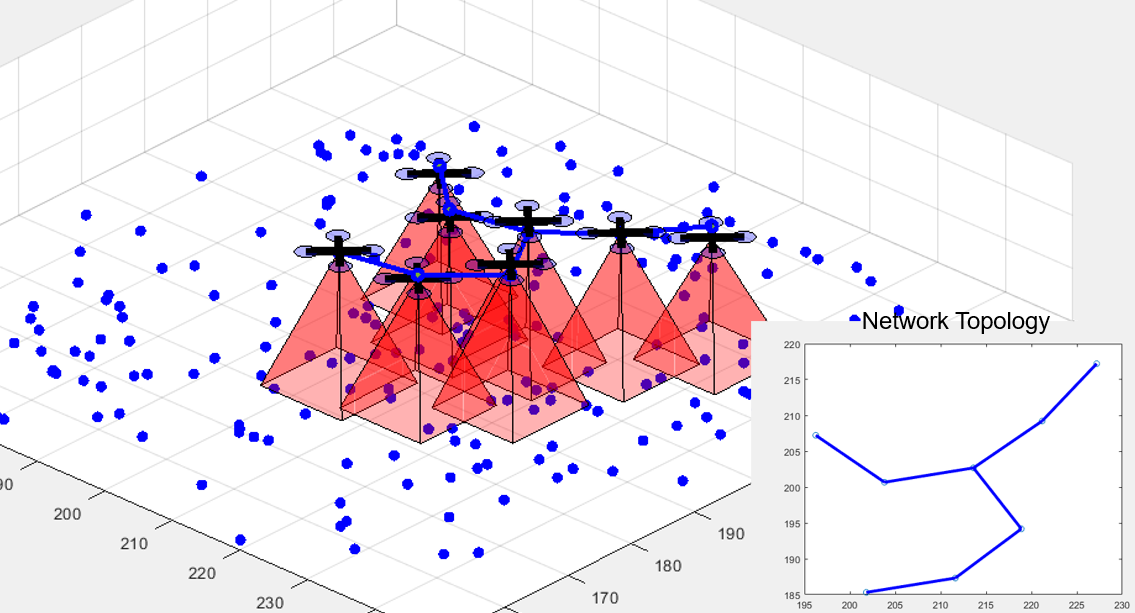}
    }
	\caption{Screenshots of target tracking process. There are eight robots tracking moving targets. The bottom right small graph in each figure show the network topology of the team at the start of that epoch. Sub-figures (a)-(c) show how the topology evolves over time.} 
	\label{fig:animation}
\end{figure*} 

\begin{enumerate}
    \item $w_i$ is set to be the individual gain of each robot, i.e., \begin{equation*}
        w_i = f(s_i),
    \end{equation*} 
    where $s_i$ is the trajectory for robot $i$ selected using greedy algorithm. We will refer this approach as \textit{Weight1} in the Sec. \ref{sec:simulation results}. The intuition here is that if a robot itself alone can observe many targets, it is less willing to deviate from its selection.
    \item $w_i$ is set to be marginal gain of robot $i$'s selected trajectory $s_i \in \mathcal{T}_i$, i.e., 
    \begin{equation*}
        w_i = f(\mathcal{S})-f(\mathcal{S} \setminus s_i),
    \end{equation*}
    where $\mathcal{S}$ is a set of trajectories selected using greedy strategy. We will refer this approach as \textit{Weight2} in the Sec. \ref{sec:simulation results}. The intuition here is that the willingness of a robot to deviate from the greedy selection is reflected in its contribution to the whole team. The more it contributes to the team, i.e., large marginal gain, the less willing it is to deviate from its selection. 
    \item Another way to set $w_i$ is to consider the variation of objective values around the point $x_i^g$. Specifically, given the position $x_i^g$ of robot $i$ corresponding to its greedy selection, we uniformly sample a set $\{x^i_1, \ldots, x^i_W \mid \norm{x^i_j - x_i^g} \leq 1, j=1,\ldots,W\}$ and define the weight for robot $i$ as 
\begin{equation*}
    w_i = f(x_i^g) - \min_{x \in \{x^i_1, \ldots, x^i_W \}}f(x).
\end{equation*}
We will refer this approach as \textit{Weight3} in the Sec. \ref{sec:simulation results}. The intuition here is that a robot will check the drop of the number of targets observed if he deviates from its selection. If the drop is large, then it is less willing to deviate from its selection.
\end{enumerate}
It should be noted that after setting $w_i$ to be some non-negative constant, Problem \ref{problem:deviation minimization} becomes a standard QP problem and can be solved using commercial solvers such as Gurobi.

\subsection{Simulation Results}\label{sec:simulation results}
All experiments were performed on a Windows 64-bit laptop with 16 GB RAM and an 8-core Intel i5-8250U 1.6GHz CPU using MATLAB with Gurobi 9.0 \cite{gurobi}, which can deal with both convex and non-convex quadratic constraints. 
We test three algorithms in this section: the proposed algorithm, the greedy algorithm without considering communication constraints, and SGG. Here the greedy algorithm corresponds to the greedy selection at the beginning of each epoch without considering communication constraints, which can be viewed as an empirical upper bound of the team performance at each epoch. We compare with such results to show that the deviation minimization part of the proposed algorithm will only slightly reduce the performance of the team from the upper bound. The data on SGG is collected in the following way: we initialize the robots and targets in the same positions as that in the test setup of the proposed algorithm and run SGG to track targets. We use the results of SGG as a baseline.

\begin{figure}[ht]
    \subfloat[\label{fig:simulation 1}]{
    \includegraphics[width=0.22\textwidth]{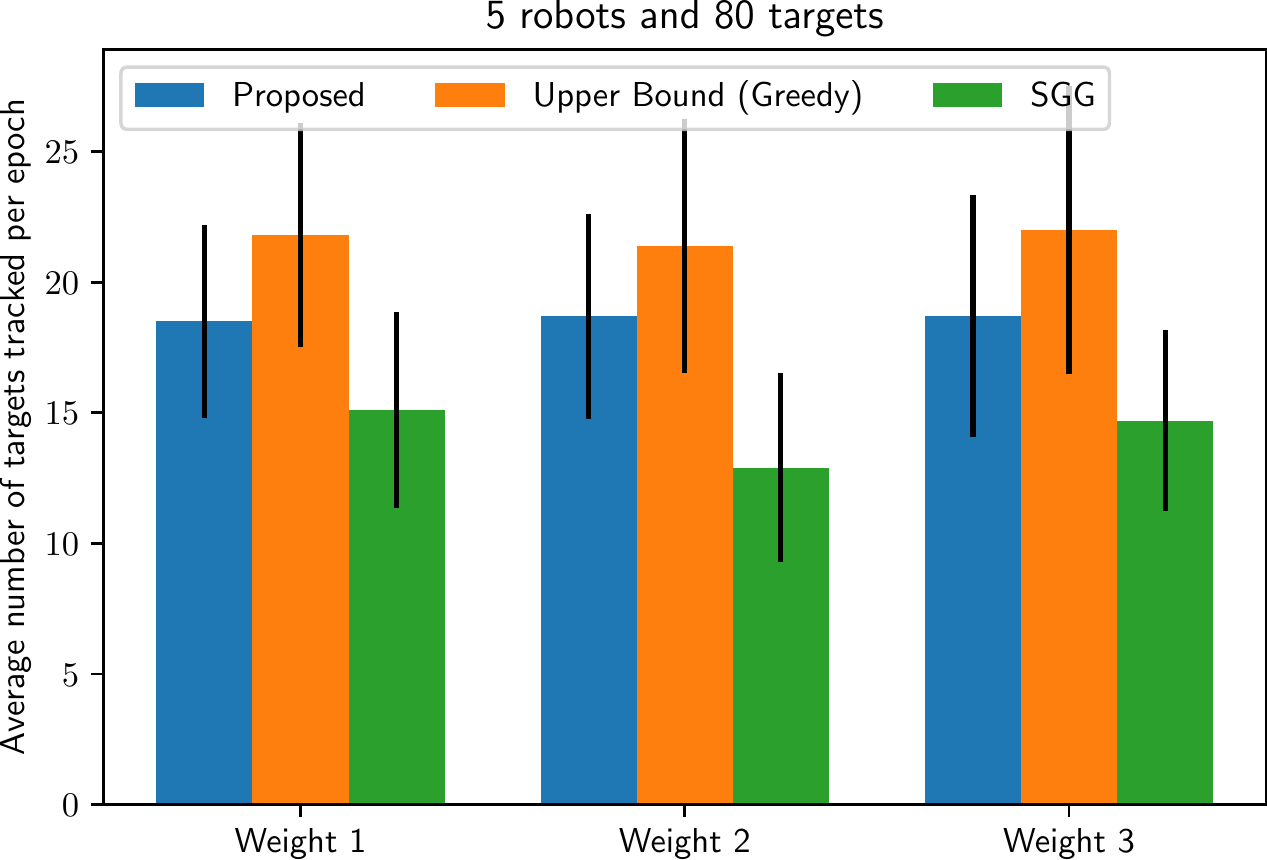}
    }
    \subfloat[\label{fig:simulation 2}]{
    \includegraphics[width=0.22\textwidth]{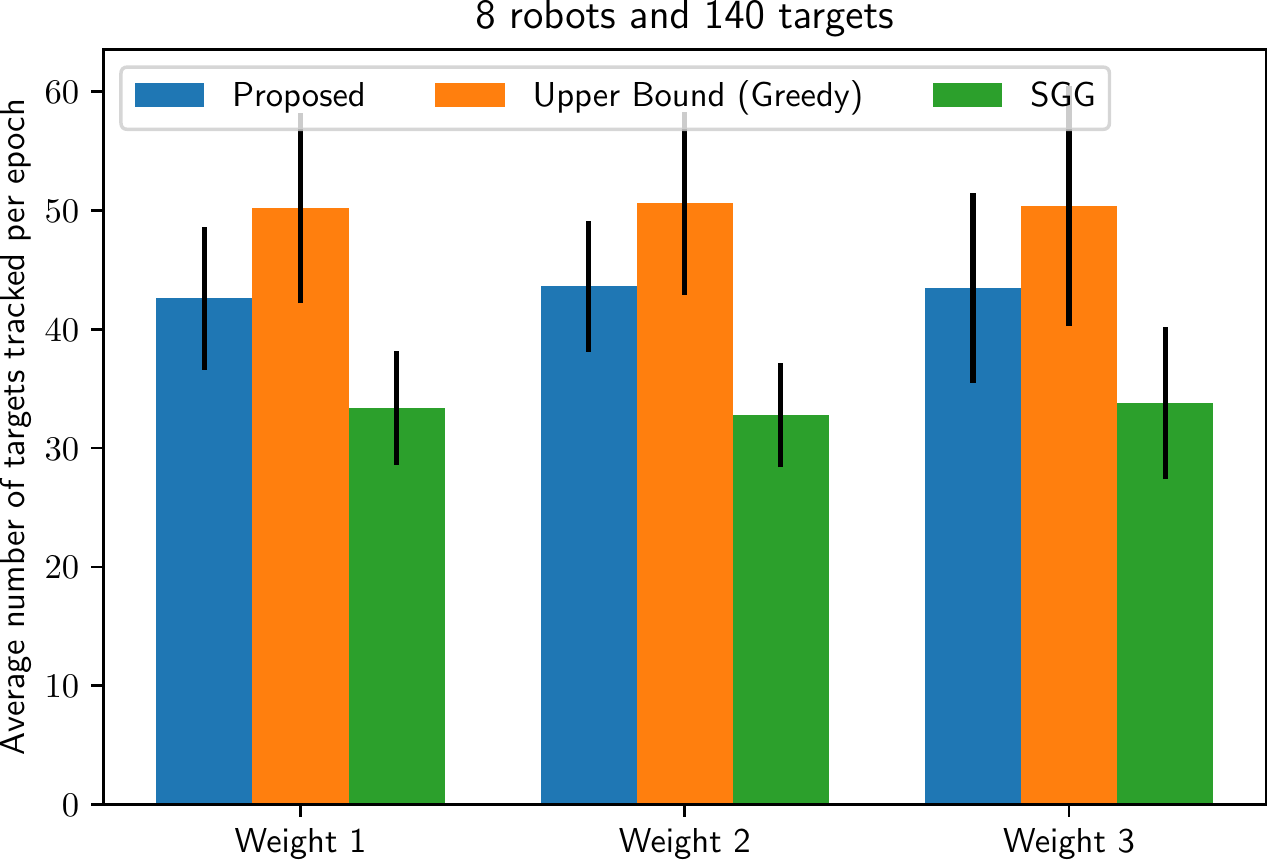}
    }\\
    \subfloat[\label{fig:simulation 3}]{
    \centerline{
    \includegraphics[width=0.22\textwidth]{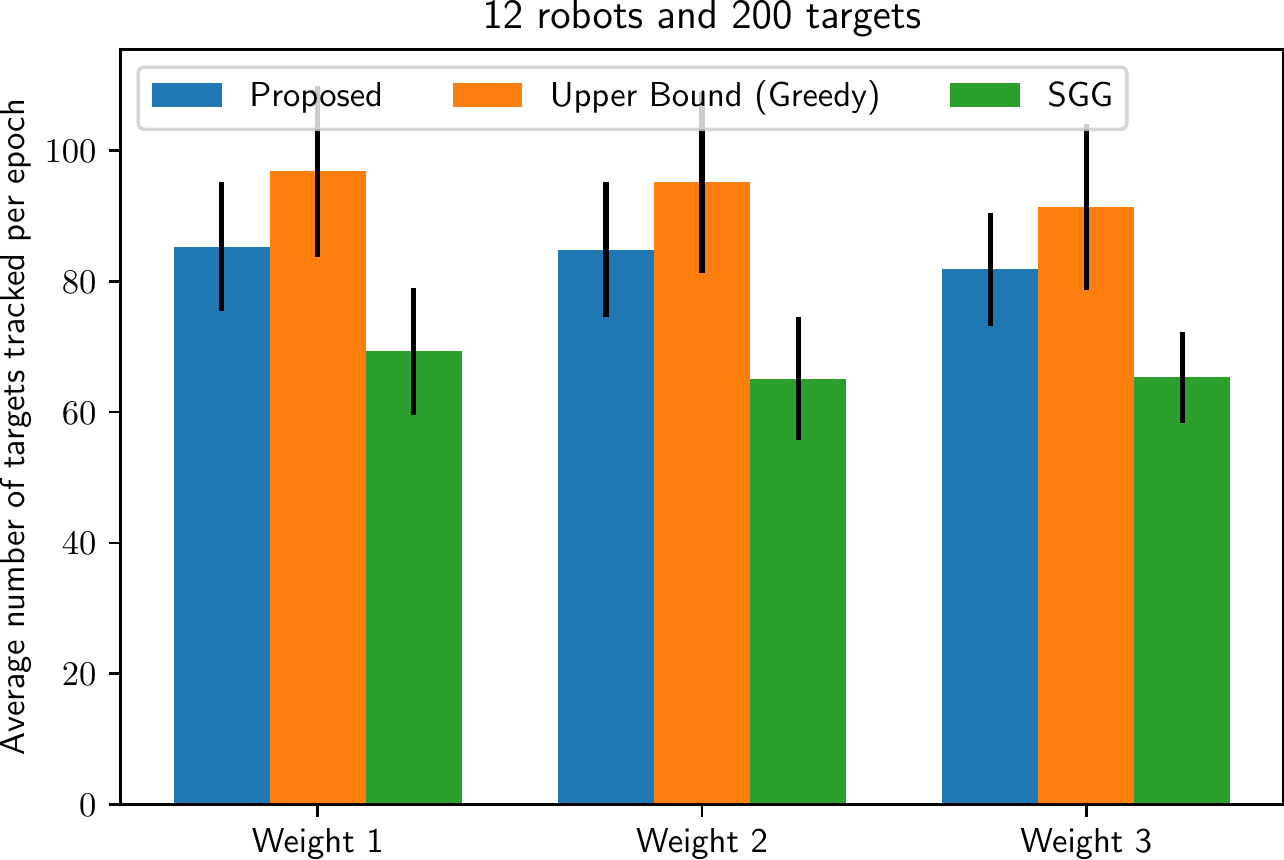}
    }
    }
	\caption{Average number of targets observed over each epoch. (a) There are 5 robots and 80 targets. (b) There are 8 robots and 140 targets. (c) There are 12 robots and 200 targets. The proposed algorithm, greedy algorithm without considering communication at each epoch, and SGG are denoted in blue, orange, and greed bars respectively. The black line above bars represent one standard deviation.} 
	\label{fig:targets_tracked}
\end{figure}

As shown in Fig. \ref{fig:targets_tracked}, the proposed algorithm can achieve the 90\% of the performance of the pure greedy strategy. Moreover, the proposed algorithm on average can track ten percent more targets compared to SGG. As for three ways to select weights in deviation minimization problem, they have similar performances. Readers are referred to multimedia submission for an animation that shows the connectivity during operation of the team and one screenshot is shown in Fig. \ref{fig:animation}.

\section{Conclusion and Future Work}
In this paper, we propose a problem named Communication-aware Submodular Maximization (CSM) for a class of multi-robot  task  planning and propose a heuristic algorithm consisting of two stages, topology generation and deviation minimization, to solve CSM. There are two directions to extend this work. One is to come up with bounded approximation algorithm to solve CSM or analyze the conditions under which some performance guarantees can be claimed. Another direction is to find efficient algorithms to approximate and discretize the reachable set in the cluttered environment. 
\section{Acknowledgment}
This work is supported by the National Science Foundation under Grant No. 1943368, and the Office of Naval Research under Grant No. N000141812829
\bibliographystyle{IEEEtran}
\bibliography{IEEEabrv,ICRA2021}

\end{document}